\documentclass{article}
\usepackage{spconf,amsmath,graphicx,subcaption,nicematrix,booktabs}

\title{M-SpeechCLIP: Leveraging Large-Scale, Pre-Trained Models for Multilingual Speech to Image Retrieval}
\name{Layne Berry$^{\text{1}}$, Yi-Jen Shih$^{\text{2}}$, Hsuan-Fu Wang$^{\text{2}}$, Heng-Jui Chang$^{\text{3}}$, Hung-yi Lee$^{\text{2}}$, and David Harwath$^{\text{1}}$}
\address{
    $^{\text{1}}$University of Texas at Austin\\
    $^{\text{2}}$National Taiwan University\\
    $^{\text{3}}$MIT CSAIL
}
\begin{document}
 \ninept

\maketitle
\begin{abstract}
This work investigates the use of large-scale, English-only pre-trained models (CLIP and HuBERT) for multilingual image-speech retrieval. 
For non-English image-speech retrieval, we outperform the current state-of-the-art performance by a wide margin both when training separate models for each language, and with a single model which processes speech in all three languages. 
We identify key differences in model behavior and performance between English and non-English settings, attributable to the English-only pre-training of CLIP and HuBERT, and investigate how fine-tuning the pre-trained models impacts these differences. 
Finally, we show that our models can be used for mono- and cross-lingual speech-text retrieval and cross-lingual speech-speech retrieval, despite never having seen any parallel speech-text or speech-speech data during training.
\end{abstract}
\section{Introduction}
\label{sec:intro}

\let\thefootnote\relax\footnotetext{This work was partially supported by JSALT 2022 workshop at Johns Hopkins University, with gift funds from Amazon, Microsoft, and Google.}

Language is more than a probability distribution over words or phonemes--it is produced in context to express semantic information.
The task of language-image retrieval targets semantic understanding by asking models to connect utterances to contexts in a different modality.
This is especially important for low-resource and unwritten languages, for which limited data exists for training speech recognition and natural language understanding systems.
The recently-proposed CLIP~\cite{CLIP} model was pre-trained on an unprecedented amount of parallel English image-text data to encode each modality in a shared semantic embedding space. SpeechCLIP~\cite{SpeechCLIP} was then proposed to map HuBERT representations of English speech into the same embedding space, achieving state-of-the-art performance on English image-speech retrieval and enabling zero-shot speech-text retrieval.
Here, we investigate the value of English-only pre-training for non-English speech understanding by applying the SpeechCLIP~\cite{SpeechCLIP} model to non-English image-speech retrieval. 
We find that our models beat the prior state-of-the-art for non-English image-speech retrieval by a wide margin.

We next train multilingual models which can take input text in any of the three languages investigated here.
We experiment with scaling up these models, and achieve further gains.
As with SpeechCLIP, we show that our models can perform zero-shot transfer to English speech-text retrieval, even outperforming image-text retrieval with the image embeddings M-SpeechCLIP used as labels during training.
Finally, we consider the challenging settings of zero-shot transfer to cross-lingual speech-text and speech-speech retrieval, setting strong baselines for the former and outperforming prior work for the latter.

This work demonstrates that large-scale pre-training is highly effective, even for tasks where both language and modality differ.
We set a new state-of-the-art for non-English image-speech retrieval, and enable both speech-text and speech-speech retrieval in a cross-lingual setting without any parallel speech from different languages, parallel speech and text, or non-English text at all.

\section{Related Work}
\label{sec:related_work}

CLIP~\cite{CLIP} is an image-text alignment model trained on 400 million web-scraped image and English caption pairs–a private dataset 4000x larger than the popular MS-COCO~\cite{MSCOCO} and Visual Genome~\cite{VisGen} and less noisy than the 4x smaller YFCC100M~\cite{YFCC100M}.
CLIP-Large consists of a ViT-L/14~\cite{ViT} image encoder and a GPT-based~\cite{GPT2} text encoder which map their respective input modalities into a shared embedding space learned with a contrastive loss. 
This embedding space has been shown to transfer well to tasks other than English image-text alignment. 
M-CLIP\cite{MCLIP} uses the CLIP text encoder with English inputs as a teacher model to learn non-English and multilingual text encoders into the CLIP embedding space, and beat state-of-the-art image-text retrieval performance on the XTD~\cite{XTD} dataset of MS-COCO caption translations in 11 languages by an average of over 12\% absolute R@10. 
SpeechCLIP~\cite{SpeechCLIP} freezes the CLIP image encoder and learns to map English spoken captions into the CLIP embedding space, beating state-of-the-art image-speech retrieval~\cite{fastvgs} on the SpokenCOCO~\cite{SpokenCOCO} and Flickr8k~\cite{Flickr8k} datasets.

Non-English image-speech retrieval has most frequently been evaluated on the Places~\cite{Places205} dataset.
\cite{PlacesEnglish} collected 400k spontaneous English spoken captions for Places images. 
\cite{PlacesHindi} collected spontaneous Hindi spoken captions for 100k of these images, and \cite{PlacesJapanese} collected spontaneous Japanese spoken captions for the same subset of images. 
\cite{PairExpansion}, \cite{Havard}, and \cite{AVLnet} also investigate non-English image-speech retrieval, with \cite{AVLnet} using transfer learning from English pre-training on HowTo100M to beat prior state-of-the-art by an average absolute R@10 of 5.7\% for Japanese and 8.1\% for Hindi.

\cite{SpeechCLIP} showed that encoding English speech into the CLIP embedding space allows speech-text retrieval to be learned without any speech-text pairs. 
Visual grounding has also been used to learn to retrieve speech given a text keyword without seeing any speech-text pairs during training. 
\cite{Sem2Speech,VT2Speech} investigate English speech retrieval given an English query word, while \cite{XKW} learns to retrieve English speech given one-word text query in German.
\cite{PlacesJapanese} used a combination of cross-lingual and cross-modal loss to learn monolingual encoders that can be used for cross-lingual speech-to-speech retrieval.

\section{Model Design}
\label{sec:method}

\subsection{Monolingual Models}
\label{ssec:method_mono}


Our model architecture is based on that of Parallel SpeechCLIP, introduced in \cite{SpeechCLIP}. 
Given a batch of $n$ image-caption pairs $(I_i,S_i)$ for $i=1, ..., n$, we use a frozen CLIP~\cite{CLIP} image encoder to generate an embedding $v^I_i$ for each image $I_i$ and train a speech encoder to produce vectors $v^S_j$ such that $v^I_i$ and $v^S_j$ are similar for $i=j$ and dissimilar for $i\neq j$.
Our speech encoder uses a frozen pre-trained HuBERT~\cite{HuBERT} model to extract frame-level audio features $f_1,...,f_k$, where $k$ is the number of frames in the audio at 50Hz.
We learn weights $\mathbf{w}=(w^1,...,w^l)$ for each of the $l$ HuBERT layers, which we use to compute the representation for each audio frame $f_t$ from the hidden states $\mathbf{h}_t=(h_t^1,...,h_t^l)$ at that frame as $f_t = \sum_{i=1}^lw^ih_t^i$. 
We then append a learnable [CLS] token to the beginning of the sequence and pass it to a trainable Transformer Encoder~\cite{Transformer}. 
The hidden state of the [CLS] token at the last layer of the Transformer Encoder on input $S_i$ is projected to the same shape as the target $v_i^I$ and used as the output encoding $v_i^S$. 
We use Masked Margin Softmax~\cite{MMS} as our contrastive loss function in both retrieval directions, so that for each batch of $B$ speech encodings $\mathbf{v}^S=(v_1^S,...,v_B^S)$ and $B$ image encodings $\mathbf{v}^I=(v_1^I,...,v_B^I)$, our total loss is $\mathcal{L}(\mathbf{v}^S,\textbf{v}^I)=L_{MMS}(\mathbf{v}^S,\mathbf{v}^I)+L_{MMS}(\mathbf{v}^I,\mathbf{v}^S)$, where $L_{MMS}$ with margin $\delta$ is defined as:
\begin{equation}
    L_{MMS}(\mathbf{x},\mathbf{y})=-\frac{1}{B}\sum_{i=1}^B\Big(\log\frac{e^{x_i\cdot y_i-\delta}}{e^{x_i\cdot y_i-\delta}+\sum_{j\neq i}^Be^{x_i\cdot y_j}}\Big)
\end{equation}

\begin{figure}[]
    \centering
    \includegraphics[width=\linewidth]{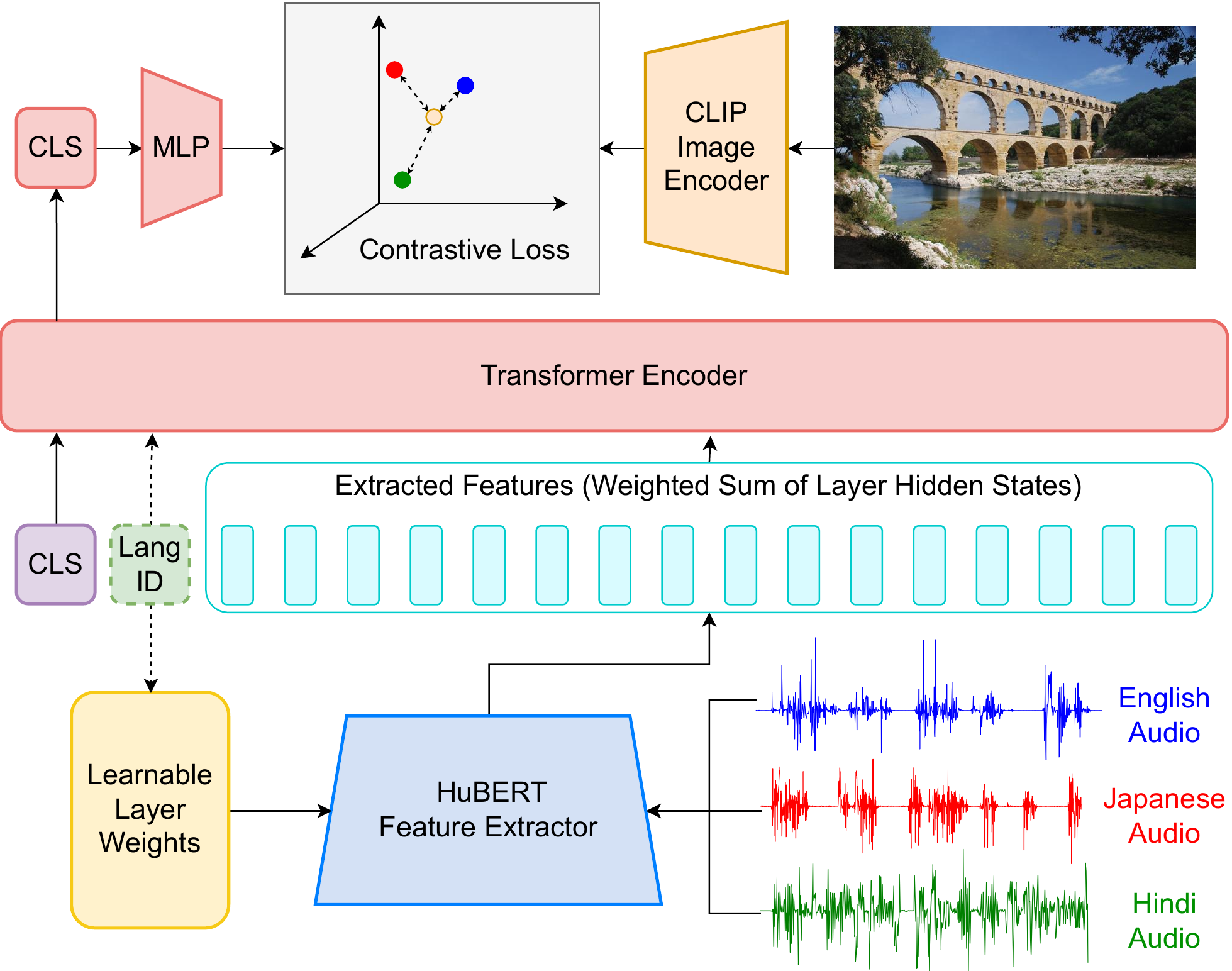}
    \caption{Diagram of the Multilingual SpeechCLIP Model}
    \label{fig:diagram}
\end{figure}

\subsection{Multilingual Models}
\label{ssec:method_multi}

We consider two possible settings: a language-agnostic model identical to the monolingual models but trained on inputs from multiple languages, and a language-aware model which learns a [LangID] token for each language which is appended to the audio feature sequence after the [CLS] token.
The language-aware model also learns a separate set of layer weights $\mathbf{w}^C$ for each language $C$.


When training multilingual models, individual batches may contain speech from just one language at a time or a mix of speech from all languages. 
Since the other pairs in a batch are used as negatives in the contrastive loss, this choice determines whether distractors are all in the same language or may be in any language. 

Finally, we consider introducing cross-lingual loss terms modeled after \cite{PlacesJapanese}. At each iteration, we randomly select two languages $C$ and $D$ and compute embeddings of the image ($\mathbf{v}^I$) and the spoken caption for that image in each selected language ($\mathbf{v}^C$ and $\mathbf{v}^D$). We then compute our total loss using six contrastive terms:
\begin{equation}
\begin{split}
    \mathcal{L}(\mathbf{v}^I,\mathbf{v}^C,\mathbf{v}^D) = &L_{MMS}(\mathbf{v}^I,\mathbf{v}^C)+L_{MMS}(\mathbf{v}^C,\mathbf{v}^I)+\\
    &L_{MMS}(\mathbf{v}^I,\mathbf{v}^D)+L_{MMS}(\mathbf{v}^D,\mathbf{v}^I)+\\
    &L_{MMS}(\mathbf{v}^C,\mathbf{v}^D)+L_{MMS}(\mathbf{v}^D,\mathbf{v}^C)
\end{split}
\end{equation}

\section{Experiments}
\label{sec:experiments}
We train and evaluate our models on Places100k, the 100k image subset of Places for which spontaneous spoken captions are available in three languages (English, Hindi, and Japanese). 
All of our monolingual models are trained on 2 V100 GPUs, as are our four multilingual models ablating batch composition and the use of a [LangID] token. 
We empirically select 1 layer and 8 attention heads for our Transformer Encoders.
Monolingual models are trained for 25 epochs and take 1 day to finish training, while multilingual models (which see 3x more training data per epoch) are trained for 15 epochs and take 1.5 days. The batch size used for these experiments is 256, and the feature extractor is HuBERT-Large.

We additionally conduct two experiments on 8 V100 GPUs with a batch size of 72, in order to test more computationally expensive ablations to our multilingual model. 
These experiments take longer to converge, so we run them for 25 epochs ($\sim$3 days).
The first, which we refer to as ``Multi+TrainFeat", unfreezes the HuBERT-Large feature extractor, allowing it to be fine-tuned on our multilingual input data. 
For the second, which we call ``Multi+TrainFeat+XLL", we include cross-lingual loss terms.
Since this doubles the number of gradients which must be stored at each training step, we use a trainable HuBERT-Base as our feature extractor for this experiment. 
This allows the amount of compute required by the ``Multi+TrainFeat" and the ``Multi+TrainFeat+XLL" models to be comparable, and the same batch size, number of GPUs, and training time to be used for both experiments.

Our learning rates increase linearly from zero to a peak over the first 10\% of iterations and decrease linearly back to zero over the remaining 90\% of iterations. 
The peak learning rate was set empirically to 0.001 for English and Japanese monolingual models and 0.0005 for Hindi monolingual and all multilingual models. 
The MMS margin $\delta$ is set to $0.001$ and spoken captions are zero-padded or truncated as necessary to 15s.

\section{Performance and Analysis}
\label{sec:results}

\subsection{Monolingual Image-Speech Retrieval}
\label{ssec:results_mono}


We evaluate the quality of the learned speech encoders by measuring their retrieval performance on the Places100k test set of 1000 image-speech pairs. 
Following prior works~\cite{PlacesHindi,PlacesJapanese,PairExpansion,AVLnet}, we report Recall@$k$ for $k=1,5,10$ in Table~\ref{tab:multi}. 
Our models' improvement over prior work is significantly larger for English than other languages. Still, for Hindi and Japanese, our models achieve an average gain of over 7\% absolute R@10 over the state-of-the-art~\cite{AVLnet}. 
Lower recall for Hindi than English and Japanese is in line with prior work; lower recall for Japanese than English, however, contrasts the findings of~\cite{PlacesJapanese} and \cite{PairExpansion}. This divergence from the trends of prior work indicates that English-only pre-training biases our models in favor of English speech.

\subsection{Multilingual Image-Speech Retrieval}
\label{ssec:results_multi}
We evaluate our multilingual models on the same image-speech retrieval task, and report results in Table \ref{tab:multi}.
When the feature extractors are frozen, monolingual models outperform multilingual ones in monolingual evaluation settings.
Our language-aware and language-agnostic models perform similarly, indicating that providing an explicit language ID to the model is not necessary. 
On the other hand, we find that using mixed-language batches during training produces much better models than those which only consider distractors in the same language.
This finding holds across both multilingual \textit{and} monolingual tests. 
We therefore select the language-agnostic architecture and multilingual training batches for our two more computationally expensive experiments.

Allowing the feature extractor to be fine-tuned leads to sizable gains both in monolingual and in mixed-language evaluation settings, particularly for non-English inputs.
Despite using only a HuBERT-Base feature extractor, the model trained with cross-lingual loss terms achieves additional gains on Hindi, Japanese, and mixed-language tests. 
Improvements from the cross-lingual loss terms are less pronounced in the monolingual English evaluation setting, making ``Multi+TrainFeat+XLL" our only model to achieve better scores on the Places100k Japanese test set than the English.

For a multilingual baseline, we consider cascading ASR with the text-based multilingual M-CLIP~\cite{MCLIP} model. 
The English and Hindi Places datasets include ASR-generated transcriptions of the spoken captions; for the Japanese data, we use the XLSR-53 Large model fine-tuned on Japanese to generate transcriptions. 
We then use the XLM-R Large ViT-L/14 variant of M-CLIP to produce embeddings of the transcribed captions, which we use for image-speech retrieval. All of our models outperform this baseline, especially in the Japanese and multilingual evaluation settings.


\begin{table}[]
    \centering
    \setlength\tabcolsep{2.5pt}
    \begin{tabular}{lcccccc}
       \toprule
         & \multicolumn{3}{c}{Image$\rightarrow$Speech} & \multicolumn{3}{c}{Speech$\rightarrow$Image} \\
         & R@1 & R@5 & R@10 & R@1 & R@5 & R@10 \\
       \midrule
        & \multicolumn{6}{c}{Japanese} \\
       Trilingual Embds.~\cite{PlacesJapanese} & 20.0 & 46.8 & 62.3 & 20.3 & 52.0 & 66.7 \\
       Pair Expansion~\cite{PairExpansion} & 16.7 & 44.3 & 57.8 & 20.1 & 49.7 & 63.9 \\
       AVLnet~\cite{AVLnet} & 24.3 & 56.6 & 70.0 & 23.5 & 57.3 & 70.4 \\
       Japanese-SpeechCLIP & \textbf{32.1} & \textbf{66.6} & \textbf{78.7} & \textbf{32.9} & \textbf{66.4} & \textbf{77.5}\\[-5pt]
       \multicolumn{7}{c}{\dotfill}\\
       Monolingual Batches & 21.1 & 50.0 & 63.6 & 22.5 & 48.5 & 62.4 \\
       Mono+LangID & 20.1 & 50.1 & 63.3 & 20.1 & 50.9 & 64.3 \\
       Multilingual Batches & \textbf{26.0} & \textbf{56.3} & \textbf{69.7} & \textbf{26.5} & \textbf{55.1} & 68.0 \\
       Multi+LangID & 24.5 & 55.4 & 69.0 & 24.6 & 54.5 & \textbf{69.8} \\[-5pt]
       \multicolumn{7}{c}{\dotfill}\\
       Multi+TrainFeat & 32.6 & 66.4 & 77.5 & 32.1 & 66.1 & 78.3 \\
       Multi+TrainFeat+XLL & \textbf{51.0} & \textbf{83.3} & \textbf{91.2} & \textbf{48.8} & \textbf{80.0} & \textbf{90.0} \\[-5pt]
       \multicolumn{7}{c}{\dotfill}\\
       ASR$\rightarrow$M-CLIP~\cite{MCLIP} & 12.0 & 35.2 & 46.3 & 22.9 & 47.4 & 58.5 \\
       \midrule
        & \multicolumn{6}{c}{Hindi} \\
       Bilingual Embds.~\cite{PlacesHindi} & 7.4 & 23.5 & 35.4 & 8.0 & 25.0 & 35.6 \\
       Trilingual Embds.~\cite{PlacesJapanese} & 10.8 & 31.3 & 41.9 & 11.2 & 31.5 & 44.5 \\
       Pair Expansion~\cite{PairExpansion} & 9.3 & 29.5 & 38.2 & 9.4 & 29.8 & 41.8 \\
       AVLnet~\cite{AVLnet} & 17.0 & 39.8 & 51.5 & 15.2 & 38.9 & 51.1 \\
       Hindi-SpeechCLIP & \textbf{21.8} & \textbf{46.5} & \textbf{58.8} & \textbf{19.1} & \textbf{42.7} & \textbf{57.3}\\[-5pt]
       \multicolumn{7}{c}{\dotfill}\\
       Monolingual Batches & 16.5 & 35.6 & 46.9 & 12.3 & 32.8 & 44.5 \\
       Mono+LangID & 17.1 & 37.1 & 49.4 & 12.5 & 35.7 & 47.5 \\
       Multilingual Batches & 17.6 & 41.3 & 52.9 & \textbf{16.7} & \textbf{39.1} & \textbf{51.2} \\
       Multi+LangID & \textbf{18.1} & \textbf{41.4} & \textbf{53.2} & 14.8 & 37.6 & 49.6 \\[-5pt]
       \multicolumn{7}{c}{\dotfill}\\
       Multi+TrainFeat & 23.5 & 52.5 & 64.7 & 20.7 & 48.4 & 62.2 \\
       Multi+TrainFeat+XLL & \textbf{35.4} & \textbf{66.1} & \textbf{76.3} & \textbf{34.6} & \textbf{64.9} & \textbf{75.3} \\[-5pt]
       \multicolumn{7}{c}{\dotfill}\\
       ASR$\rightarrow$M-CLIP~\cite{MCLIP} & 13.8 & 30.2 & 42.0 & 19.5 & 40.8 & 50.8 \\
       \midrule
        & \multicolumn{6}{c}{English} \\
       Bilingual Embds.~\cite{PlacesHindi} & 8.0 & 25.2 & 36.5 & 8.3 & 28.2 & 42.4 \\
       Trilingual Embds.~\cite{PlacesJapanese} & 11.6 & 35.8 & 50.8 & 13.9 & 39.5 & 52.9 \\
       Pair Expansion~\cite{PairExpansion} & 12.3 & 35.3 & 47.7 & 13.8 & 40.2 & 51.6 \\
       English-SpeechCLIP & \textbf{50.9} & \textbf{82.8} & \textbf{90.6} & \textbf{48.9} & \textbf{82.3} & \textbf{89.4} \\[-5pt]
       \multicolumn{7}{c}{\dotfill}\\
       Monolingual Batches & 37.5 & 73.2 & 84.7 & 36.1 & 70.9 & 83.1 \\
       Mono+LangID & 37.8 & 72.9 & 84.0 & 36.1 & 71.3 & 83.9 \\
       Multilingual Batches & \textbf{41.9} & 75.6 & 86.4 & \textbf{41.1} & 73.9 & 85.1 \\
       Multi+LangID & 41.1 & \textbf{77.2} & \textbf{86.5} & 40.8 & \textbf{75.0} & \textbf{86.4} \\[-5pt]
       \multicolumn{7}{c}{\dotfill}\\
       Multi+TrainFeat & 44.6 & 79.8 & \textbf{89.1} & 44.2 & 78.3 & 87.3 \\
       Multi+TrainFeat+XLL & \textbf{51.7} & \textbf{82.2} & 88.9 & \textbf{48.0} & \textbf{79.7} & \textbf{88.0} \\[-5pt]
       \multicolumn{7}{c}{\dotfill}\\
       ASR$\rightarrow$M-CLIP~\cite{MCLIP} & 29.2 & 58.1 & 69.4 & 43.4 & 70.5 & 78.9 \\
       \midrule
        & \multicolumn{6}{c}{Multilingual} \\
       Monolingual Batches & 25.0 & 55.3 & 66.8 & 29.7 & 53.5 & 65.4 \\
       Mono+LangID & 25.0 & 52.0 & 65.3 & 28.3 & 52.4 & 66.2 \\
       Multilingual Batches & \textbf{25.8} & 56.6 & 68.4 & \textbf{33.1} & \textbf{56.2} & 66.7 \\
       Multi+LangID & 24.9 & \textbf{58.8} & \textbf{69.2} & 32.8 & 55.4 & \textbf{68.7} \\[-5pt]
       \multicolumn{7}{c}{\dotfill}\\
       Multi+TrainFeat & 34.0 & 63.1 & 75.9 & 31.0 & 61.5 & 74.1 \\
       Multi+TrainFeat+XLL & \textbf{44.8} & \textbf{76.3} & \textbf{85.9} & \textbf{43.0} & \textbf{74.5} & \textbf{84.9} \\[-5pt]
       \multicolumn{7}{c}{\dotfill}\\
       ASR$\rightarrow$M-CLIP~\cite{MCLIP} & 20.0 & 35.7 & 45.0 & 28.5 & 52.9 & 62.3 \\
       \bottomrule
    \end{tabular}
    \caption{Retrieval performance on the Places100k test set. For each language, we evaluate (from top to bottom): monolingual models; multilingual models trained on 2 V100 GPUs; multilingual models trained on 8 V100 GPUs; and monolingual ASR cascaded with multilingual image-text retrieval from prior work. All but the monolingual models are also evaluated on a multilingual test set formed by randomly selecting a language for each image's spoken caption.
    }
    \label{tab:multi}
\end{table}

\subsection{Analysis of Learned Layer Weights}

\begin{figure}[]
    \centering
    \begin{subfigure}{\columnwidth}
        \includegraphics[width=\columnwidth]{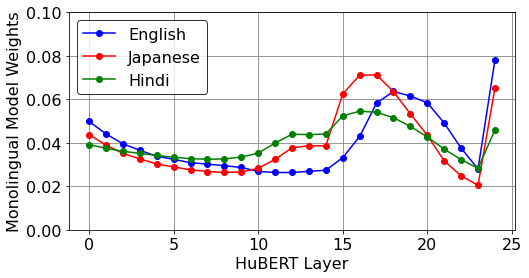}
    \end{subfigure}
    
    \begin{subfigure}{\columnwidth}
        \includegraphics[width=\columnwidth]{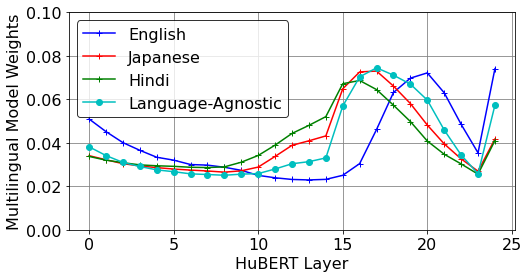}
    \end{subfigure}
    \caption{Top: Layer weights learned by monolingual models. Bottom: Layer weights learned by multilingual models trained on multilingual batches with a frozen HuBERT-Large feature extractor.}
    \label{fig:weights}
\end{figure}

Fig.~\ref{fig:weights} summarizes the layer weights learned by our models with frozen feature extractors. 
Two separate patterns emerge for English and non-English weights, regardless of whether they are learned by a monolingual or a multilingual model.
Layers 15 through 18 receive the highest weights when processing non-English inputs, whereas layers 17 through 21 receive the highest weights when processing English speech, suggesting that later HuBERT layers are more specialized to English than earlier ones.
The maximum and minimum weights assigned to any layer by the language-aware multilingual model are more similar across languages than the weights learned by monolingual models.
For the language-agnostic multilingual model, which uses the same weights for both English and non-English speech, layer weights are distributed according to the non-English pattern.
This does not appear to hurt the model's retrieval performance on English-only test sets, which Table~\ref{tab:multi} shows is comparable for the language-agnostic and language-aware models.

\subsection{Zero-Shot Speech-Text Retrieval}
\label{ssec:s2t_t2s}

Since our targets $\mathbf{v}^I$ are image embeddings in the output space of CLIP-Large, we can compare our speech embeddings $\mathbf{v}^S$ to the output $\mathbf{v}^T$ of the CLIP-Large text encoder on the English captions for each image in the batch. 
We report the performance of our monolingual models and our two large multilingual models on this task.

Inspired by~\cite{Sem2Speech}, we consider image-text retrieval using the CLIP image and text encoders as a point of comparison. 
Since the CLIP image encoder outputs were used as our targets during training, we would expect this to function as a topline--the performance our model would achieve if it output precisely the training targets. 
Instead, as in~\cite{Sem2Speech}, our models significantly outperform this experiment, with improvements of nearly 15\% when English text is the query and over 20\% when English text is being retrieved.

In the cross-lingual setting, despite the difficulty of the task, our models achieve reasonable performance. 
Our monolingual Japanese model performs the best in the Japanese Speech$\rightarrow$English Text direction, while the multilingual model trained with cross-lingual loss performs best for all other directions.
This is the first work to successfully perform zero-shot transfer from image-speech retrieval to cross-lingual text-speech retrieval with texts longer than a single word, and sets a strong baseline for the task. 
We achieve this with no text at all in the spoken source languages. 

\begin{table}[t]
    \centering
    \setlength\tabcolsep{2.5pt}
    \begin{tabular}{lcccccc}
       \toprule
         & \multicolumn{3}{c}{Prompt$\rightarrow$Target} & \multicolumn{3}{c}{Target$\leftarrow$Prompt}\\
         & R@1 & R@5 & R@10 & R@1 & R@5 & R@10 \\
       \midrule
        & \multicolumn{6}{c}{Image $\leftrightarrow$ English Text} \\
       CLIP~\cite{CLIP} & 26.1 & 51.3 & 63.8 & 46.4 & 71.8 & 78.4 \\
    
        & \multicolumn{6}{c}{English Speech $\leftrightarrow$ English Text} \\
       Monolingual* & \textbf{57.1} & \textbf{80.0} & \textbf{85.1} & 67.8 & 89.1 & 93.1 \\
       Multi+TrainFeat* & 50.0 & 75.7 & 84.0 & 63.0 & 85.2 & 90.1 \\
       Multi+TrainFeat+XLL* & 39.0 & 68.7 & 79.1 & \textbf{70.1} & \textbf{90.1} & \textbf{93.2} \\
        & \multicolumn{6}{c}{Japanese Speech $\leftrightarrow$ English Text} \\
       Monolingual* & \textbf{11.8} & \textbf{34.8} & \textbf{45.1} & 12.9 & 31.3 & 42.9 \\
       Multi+TrainFeat* & 9.8 & 28.6 & 39.9 & 11.3 & 27.9 & 38.1 \\
       Multi+TrainFeat+XLL* & 10.8 & 27.7 & 39.1 & \textbf{23.9} & \textbf{48.8} & \textbf{60.1} \\
        & \multicolumn{6}{c}{Hindi Speech $\leftrightarrow$ English Text} \\
       Monolingual* & 8.4 & 22.3 & 31.7 & 8.0 & 22.5 & 32.3 \\
       Multi+TrainFeat* & 8.6 & 25.2 & 35.1 & 8.1 & 25.9 & 39.6 \\
       Multi+TrainFeat+XLL* & \textbf{14.8} & \textbf{34.5} & \textbf{45.7} & \textbf{16.9} & \textbf{32.3} & \textbf{40.8} \\
        & \multicolumn{6}{c}{English Speech $\leftrightarrow$ Japanese Speech} \\
       Trilingual Embds.~\cite{PlacesJapanese} & 10.5 & 31.2 & 43.7 & 10.6 & 31.7 & 44.1 \\
       Multi+TrainFeat* & 10.2 & 28.9 & 41.1 & 10.7 & 30.0 & 42.4 \\
       Multi+TrainFeat+XLL & \textbf{28.9} & \textbf{56.4} & \textbf{69.6} & \textbf{28.6} & \textbf{57.2} & \textbf{69.6} \\
        & \multicolumn{6}{c}{English Speech $\leftrightarrow$ Hindi Speech} \\
       Trilingual Embds.~\cite{PlacesJapanese} & 7.6 & 22.5 & 31.3 & 7.6 & 22.5 & 31.3 \\
       Multi+TrainFeat* & 9.3 & 27.6 & 40.2 & 10.1 & 28.6 & 40.6 \\
       Multi+TrainFeat+XLL & \textbf{26.0} & \textbf{49.5} & \textbf{60.2} & \textbf{25.1} & \textbf{50.9} & \textbf{61.8} \\
        & \multicolumn{6}{c}{Japanese Speech $\leftrightarrow$ Hindi Speech} \\
       Trilingual Embds.~\cite{PlacesJapanese} & 10.4 & 24.6 & 35.0 & 8.5 & 24.8 & 33.4 \\
       Multi+TrainFeat* & 7.0 & 19.4 & 29.4 & 7.7 & 18.4 & 26.4 \\
       Multi+TrainFeat+XLL & \textbf{22.5} & \textbf{46.1} & \textbf{56.3} & \textbf{22.0} & \textbf{45.8} & \textbf{59.0} \\
       \bottomrule
    \end{tabular}
    \caption{Image-text, speech-text, and speech-speech retrieval on the Places100k test set. * indicates a model is not trained on this task.
    }
    \label{tab:s2t}
\end{table}

\subsection{Cross-Lingual Speech-Speech Retrieval}
\label{ssec:multi_s2s}

Finally, we consider the task of cross-lingual speech-to-speech retrieval.
We report results for our ``Multi+TrainFeat" model, for which this is a zero-shot transfer task, as well as for our ``Multi+TrainFeat+XLL" model, for which it is not.
The monolingual models trained simultaneously with cross-lingual loss terms from~\cite{PlacesJapanese} provide a baseline for this task. 
Our multilingual model trained with the cross-lingual loss terms from~\cite{PlacesJapanese} outperforms prior work by 20-30\% R@10 for all retrieval directions. 
Even our model trained without any cross-lingual supervision can perform zero-shot transferreasonably well.
While it underperforms compared to~\cite{PlacesJapanese} in the English $\leftrightarrow$ Japanese and Japanese $\leftrightarrow$ Hindi directions, it outperforms prior work for the English $\leftrightarrow$ Hindi directions, despite the significant disadvantage of having not been trained for cross-lingual speech-speech retrieval.
This indicates that learning to align spoken captions with CLIP image embeddings can produce high-quality alignments between speech in different languages. 



\section{Conclusion}
\label{sec:conclusion}
In this paper, we found that large-scale, English-only pre-training is effective not only for downstream tasks processing English speech, but can be used to achieve state-of-the-art performance even for non-English image-speech retrieval. 
The CLIP semantic embedding space can be used to represent not only multiple modalities, but multiple languages effectively. 
Examining the layer weights learned by our model revealed that pre-trained HuBERT speech encoders specialize for English in later layers, but that features extracted from their middle layers are useful for non-English downstream tasks even without any fine-tuning. 
We also trained a single model to encode speech in multiple languages into the CLIP semantic embedding space.
This model can then be used to perform zero-shot cross-lingual speech-to-speech retrieval. 
Our models significantly outperform prior work when trained with a cross-lingual objective, and perform comparably to prior work even when trained without one. 
Finally, we showed that our learned speech encoders can perform zero-shot speech-text retrieval for English text even when the speech is not in English. 
In our future work, we plan to explore the use of these models to bootstrap speech-to-text and speech-to-speech translation for low-resource languages when neither parallel data nor non-English text is available.

\vfill\pagebreak


\bibliographystyle{IEEEbib}
\bibliography{main}

\end{document}